\documentclass{article}

     \usepackage{arxiv}

\usepackage[utf8]{inputenc} 
\usepackage[T1]{fontenc}    
\usepackage{hyperref}       
\usepackage{url}            
\usepackage{booktabs}       
\usepackage{amsfonts}       
\usepackage{nicefrac}       
\usepackage{microtype}      
\usepackage{graphicx}
\usepackage{algorithm}
\usepackage{algorithmic}
\usepackage{esvect}
\usepackage{amsthm}
\usepackage{booktabs}
\usepackage{subcaption}

\title{Explainable AI without Interpretable Model}

\author{%
  Kary Fr\"amling \\
  Department of Computing Science\\
  Ume\r{a} University\\
  Mit-huset, 901 87 Ume\r{a}, Sweden \\
  \texttt{kary.framling@umu.se} \\
}

\newtheorem{definition}{Definition}

\begin{document}

\maketitle

\begin{abstract}
Explainability has been a challenge in AI for as long as AI has existed. With the recently increased use of AI in society, it has become more important than ever that AI systems would be able to explain the reasoning behind their results also to end-users in situations such as being eliminated from a recruitment process or having a bank loan application refused by an AI system. Especially if the AI system has been trained using Machine Learning, it tends to contain too many parameters for them to be analysed and understood, which has caused them to be called `black-box' systems. Most Explainable AI (XAI) methods are based on extracting an \textit{interpretable} model that can be used for producing explanations. However, the interpretable model does not necessarily map accurately to the original black-box model. Furthermore, the understandability of interpretable models for an end-user remains questionable. The notions of Contextual Importance and Utility (CIU) presented in this paper make it possible to produce human-like explanations of black-box outcomes directly, without creating an interpretable model. Therefore, CIU explanations map accurately to the black-box model itself. CIU is completely model-agnostic and can be used with any black-box system. In addition to feature importance, the utility concept that is well-known in Decision Theory provides a new dimension to explanations compared to most existing XAI methods. Finally, CIU can produce explanations at any level of abstraction and using different vocabularies and other means of interaction, which makes it possible to adjust explanations and interaction according to the context and to the target users. 
\end{abstract}

\section{Introduction}\label{Sec:Intro}

Explainability has been a challenge in AI for as long as AI has existed. Shortliffe et al pointed out already in 1975 that `It is our belief, therefore, that a consultation program will gain acceptance only if it serves to augment rather than replace the physician’s own decision making processes.' \cite{SHORTLIFFE1975303}. The system described in that paper was MYCIN, an expert system that was capable of advising physicians who request advice regarding selection of appropriate antimicrobial therapy for hospital patients with bacterial infections. Great emphasis was put into the interaction with the end-user, in this case a skilled physician. 

With the recently increased use of AI in society, it has become more important than ever that AI systems should be able to explain the reasoning behind their results also to end-users, in situations such as being eliminated from a recruitment process or having a bank loan application refused. Meanwhile, many XAI researchers have pointed out that it is rare that current XAI research would truly take `normal' end-users into consideration. For instance, Miller et al. illustrate the phenomenon in their article entitled `Explainable AI: Beware of Inmates Running the Asylum' for expressing the tendency that current XAI methods mainly help AI researchers to understand their own results and models \cite{miller2017inmates}.

Many XAI researchers also point out that it is fair to say that most XAI work uses only the researchers' intuition of what constitutes a `good' explanation, while ignoring the vast and valuable bodies of research in philosophy, psychology, and cognitive science of how people define, generate, select, evaluate, and present explanations \cite{miller2018insights}\cite{Westberg_XAI_2019}. Another domain that seems neglected in current XAI work is \textit{Decision Theory} and related sub-domains such as \textit{Multiple Criteria Decision Making (MCDM)} \cite{KeeneyRaiffa_1976}. Decision Theory is tightly connected with the other mentioned domains because methods of Decision Theory are intended to produce Decision Support Systems (DSS) that are understood and used by humans when taking decisions. Decision Theory and MCDM provide clear definitions of what is meant by the \textit{importance} of an input, as well as what is the \textit{utility} of a given input value towards the outcome of the DSS. A simple linear DSS model is the weighted sum, where a numerical weight expresses the importance of an input and a numerical score expresses the utility of different possible input values for different outcomes of the DSS, i.e. how good or favorable a value is. 

\textit{Contextual Importance and Utility (CIU)} extends this linear definition of importance and utility towards non-linear models such as those produced by typical ML methods. In many (or most) real-life situations the importance of an input and the utility of different input values changes depending on values of other inputs. For instance, the outdoor temperature has a great importance on a person's comfort level as long as the person is outdoors. When the person goes inside, the situation (context) changes and the outdoor temperature then only has an indirect (if any) importance for the person's comfort level. Regarding utility, both a very cold and a very warm outdoor temperature might be good or bad depending on the context. For instance, a $-20^{\circ}$C temperature tends to be uncomfortable if wearing a T-shirt, whereas a $20^{\circ}$C temperature is uncomfortable if wearing winter clothes. The utility of different temperature values changes when adding or removing clothes, and vice versa, the utility of different clothes changes when the temperature changes.

After this Introduction, Section \ref{Sec:Background} goes through the most relevant state-of-the-art of XAI methods. Section \ref{Sec:CIU} presents the formal definition of CIU. Experimental results are shown in Section \ref{Sec:Experiments}. Open questions and future research are presented in Section \ref{Sec:Future}, followed by conclusions in Section \ref{Sec:Conclusions}. 

\section{Background}\label{Sec:Background}


There does not seem to be a clear agreement in XAI literature on the meaning of the terms \textit{interpretable} versus \textit{explainable}. For the rest of this paper, \textit{interpretable model} will be used to signify models whose behaviour humans can understand to some extent, such as rules or linear models. \textit{Explanation} will be used to signify what is actually presented to a user for a specific prediction or outcome. 

XAI methods can be classified into categories \textit{model explanation}, \textit{outcome explanation} and \textit{model inspection} according to \cite{GuidottiEtAl_2018}. Model explanation signifies providing a \textbf{global} explanation of the black-box model through an interpretable and transparent model. This model should be able to mimic the entire behavior of the black-box and it should also be understandable by humans. Rule extraction methods and estimation of global feature importance are examples of model explanation methods, as well as decision tree, attention model, etc. 

Outcome explanation consists in providing an explanation of the outcome of the black-box for a specific instance (or context) and can therefore be considered \textbf{local}. It is not required to explain the underlying logic of the entire black-box but only the reason for the outcome on a specific input instance. Model inspection is not truly a XAI category, it mainly refers to how model or outcome explanations are presented to users (visual or textual for instance) for understanding the black-box model or its outcome. 

Most (or all) current outcome explanation methods are so-called post-hoc methods, i.e. they require creating an intermediate interpretable model to provide explanations. The \textit{Local Interpretable Model-agnostic Explanations} (LIME) method presented in 2016 \cite{ribeiro2016i} might be considered a cornerstone regarding post-hoc outcome explanation. LIME belongs to the family of \textit{additive feature attribution methods} \cite{NIPS2017_Lundberg_XAI} that are based on the assumption that a locally linear model that represents the gradient around the current context is sufficient for outcome explanation purposes. Other methods that belong to the same family are for instance Shapley values, DeepLIFT and Layer-Wise Relevance Propagation \cite{NIPS2017_Lundberg_XAI}. 


A major challenge of all  methods that use an intermediate interpretable model (the `explanation model' in \cite{NIPS2017_Lundberg_XAI}) is to what extent the interpretable model actually corresponds to the black-box model. A rising concern among XAI researchers is that current XAI methods themselves tend to be black-boxes whose behaviour is as difficult to understand as that of the explained AI black-boxes, which causes challenges to assess to what extent XAI explanations can be trusted. Furthermore, it is not evident whether a gradient-based, locally linear model is adequate or accurate for interpreting or explaining black-box behaviour. CIU differs radically from the existing state-of-the-art in XAI because CIU does not create or use an intermediate interpretable model. 

\section{Contextual Importance and Utility (CIU)}\label{Sec:CIU}

The underlying idea behind CIU is to use a similar approach to explanation as humans do when explaining or justifying a decision to other humans. In a XAI context, the \textit{explainer} is a (X)AI system that justifies or explains its decisions or actions and the \textit{explainee} is a human (one or many) that is the target of the explanation \cite{miller2018insights}. Human explainers tend to identify what were the most important aspects that influenced their decision and start their explanation with them. Human explainers also adapt the abstraction level and vocabulary used in the explanation to their expectations about what is best understood and accepted by the explainee. It is generally not enough to explain only the taken decision, it is also often necessary to justify why another decision wasn't taken instead. 

\begin{figure}[b]
\centering
\begin{subfigure}{.3\textwidth}
\centering
\includegraphics[trim={3cm 1cm 2cm 1cm},clip,width=\textwidth]{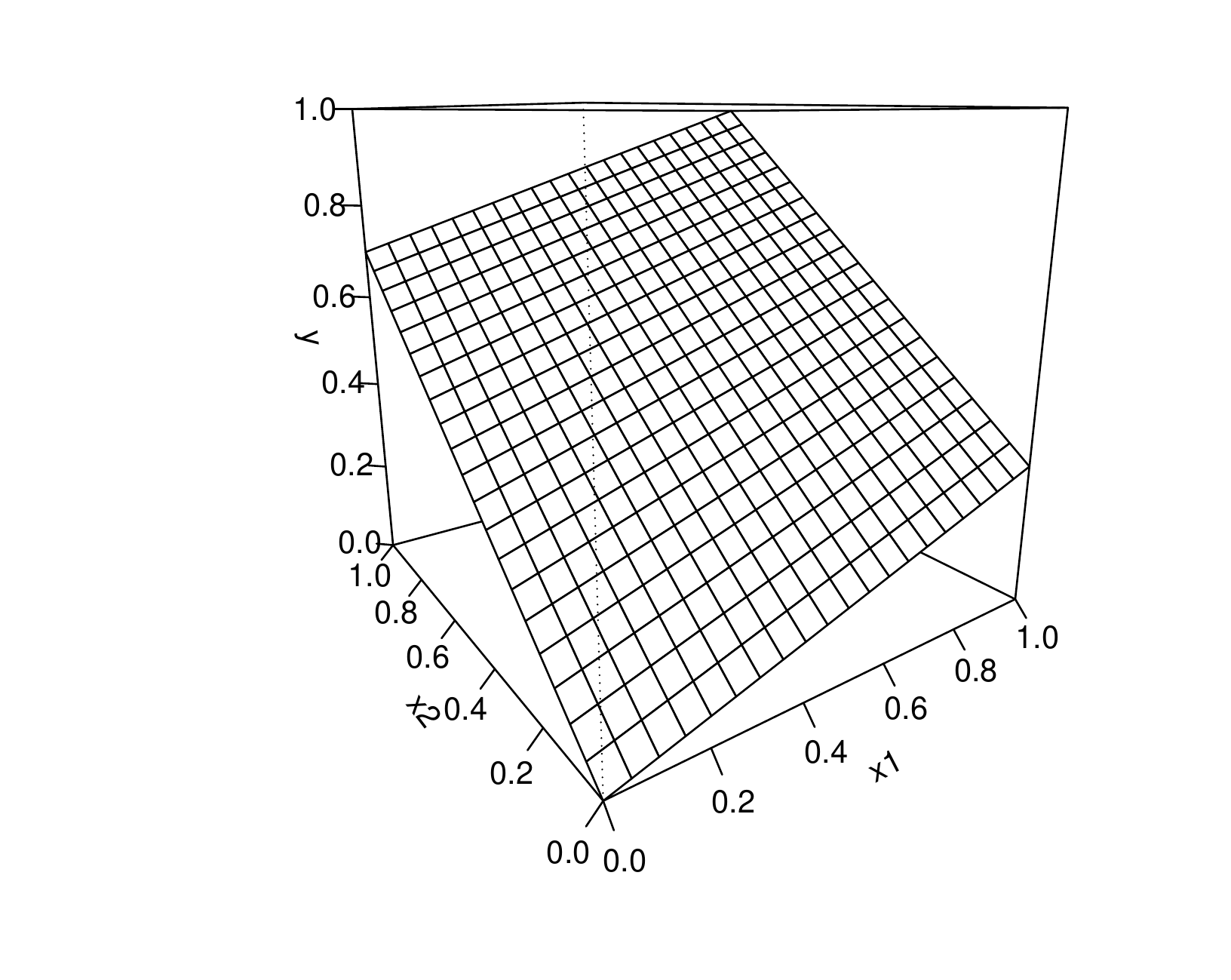}
\caption{Weighted sum.} \label{Fig:linear_func}
\end{subfigure}
\begin{subfigure}{.3\textwidth}
\centering
\includegraphics[trim={3cm 1cm 2cm 1cm},clip,width=\textwidth]{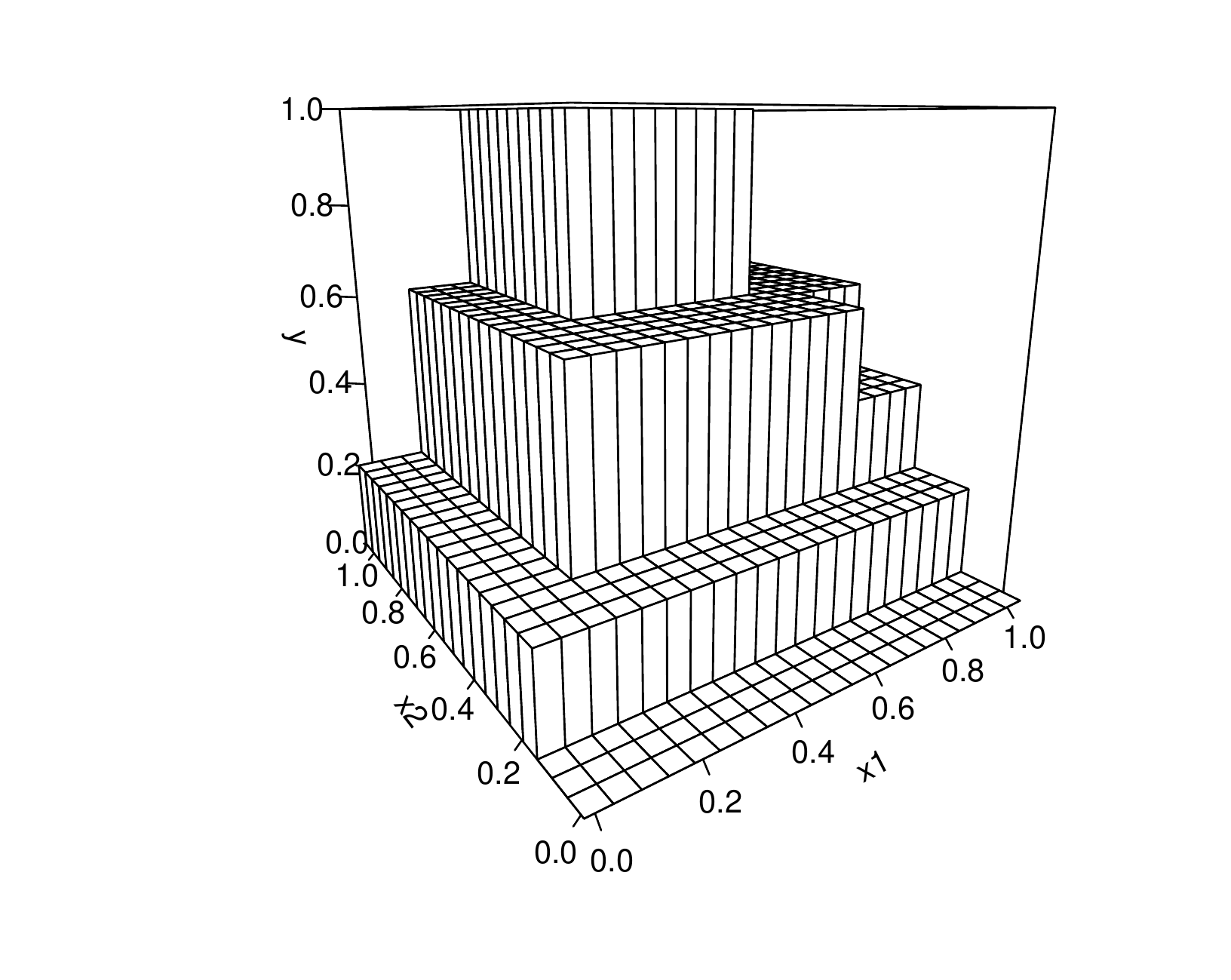}
\caption{Rule-based.} \label{Fig:rule_func}
\end{subfigure}
\begin{subfigure}{.3\textwidth}
\centering
\includegraphics[trim={3cm 1cm 2cm 1cm},clip,width=\textwidth]{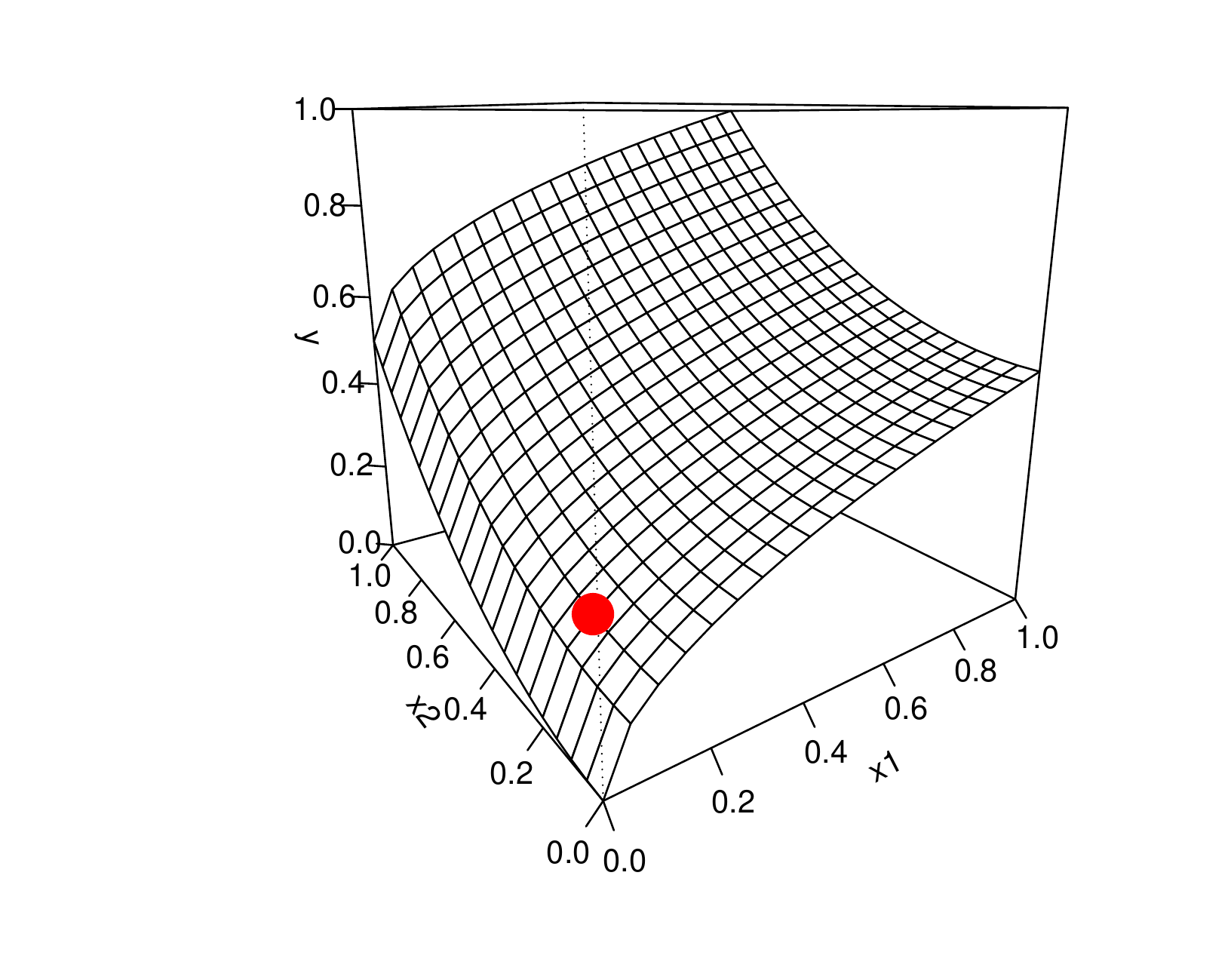}
\caption{Non-linear model.} \label{Fig:nonlinear_func}
\end{subfigure}
\caption{Examples of linear ($y = 0.3x_1 + 0.7x_2$), rule-based  and non-linear ($y = (x_{1}^{0.5} + x_{2}^2)/2$) models.}
\label{Fig:mcdm_models}
\end{figure}


CIU was initially developed in a MCDM context \cite{FramlingThesis_1996}. In MCDM, \textit{importance} and \textit{utility} concepts are clearly defined. The Analytic Hierarchy Process (AHP) \cite{AnalyticHierarchyProcess_1999} that was originally developed in the 1970's seems to have become the most popular MCDM method in research and practice \cite{KublerEtAl_ESwA_2016}. AHP is essentially based on a weighted sum, where the global output can be broken into intermediate concepts in a hierarchical manner. The importance of different criteria (features, inputs of the model) is expressed by numeric weights. The utility expresses how good, favorable or typical a value is for the output of the model. For a car selection problem, importance and utility can be used for giving explanations such as `This car is \underline{good} because it has a \underline{good} size, \underline{decent} performances and a \underline{reasonable} price, which are very important features', where words indicating utilities are underlined and only the most important features are presented. The use of a linear model makes the meaning of importance and utility quite understandable to humans, as illustrated in Figure \ref{Fig:linear_func}. 

Rule-based systems, as well as classification trees for instance, are a way of overcoming the linearity limitation but tends to lead to step-wise models as illustrated in Figure \ref{Fig:rule_func}. Non-linear models such as neural nets can learn smooth and non-linear functions as illustrated in Figure \ref{Fig:nonlinear_func}. Even though CIU can deal with all three kinds of models, the focus here is on the kind of non-linear functions in Figure \ref{Fig:nonlinear_func}. We will begin the formal definition of CIU by providing a set of definitions. 

\begin{definition}[Black-box model]
A black-box model is a mathematical transformation $f$ that maps inputs $\vv{x}$ to outputs $\vv{y}$ according to $\vv{y}=f(\vv{x})$. 
\end{definition}

\begin{definition}[Context]
A Context $\vv{C}$ defines the input values $\vv{x}$ that describe the current situation or instance to be explained.
\end{definition}

\begin{definition}[Pre-defined output range]
The value range $[absmin_{j},absmax_{j}]$ that an output $y_{j}$ can take by definition. 
\end{definition}

In classification tasks, the Pre-defined output range is typically $[0,1]$. In regression tasks the minimum and maximum output values present in a training set used for Machine Learning can usually be used as an estimate of $[absmin_{j},absmax_{j}]$. 

\begin{definition}[Set of studied inputs for CIU]
The index set $\{i\}$ defines the indices of inputs $\vv{x}$ for which CIU is calculated.
\end{definition}

\begin{definition}[Estimated output range]
$[Cmin_{j}(\vv{C},\{i\}),Cmax_{j}(\vv{C},\{i\})]$ is the range of values that an output $y_{j}$ can take in the Context $\vv{C}$ when modifying the values of inputs $x_{\{i\}}$.
\label{Def:EstimatedOutputRange}
\end{definition}

The values used for the inputs $x_{\{i\}}$ should be `representative' or realistic within the Context $\vv{C}$. The meaning of `representative' is discussed further down in this paper. 

\begin{figure}
\centering
\includegraphics[width=0.9\textwidth]{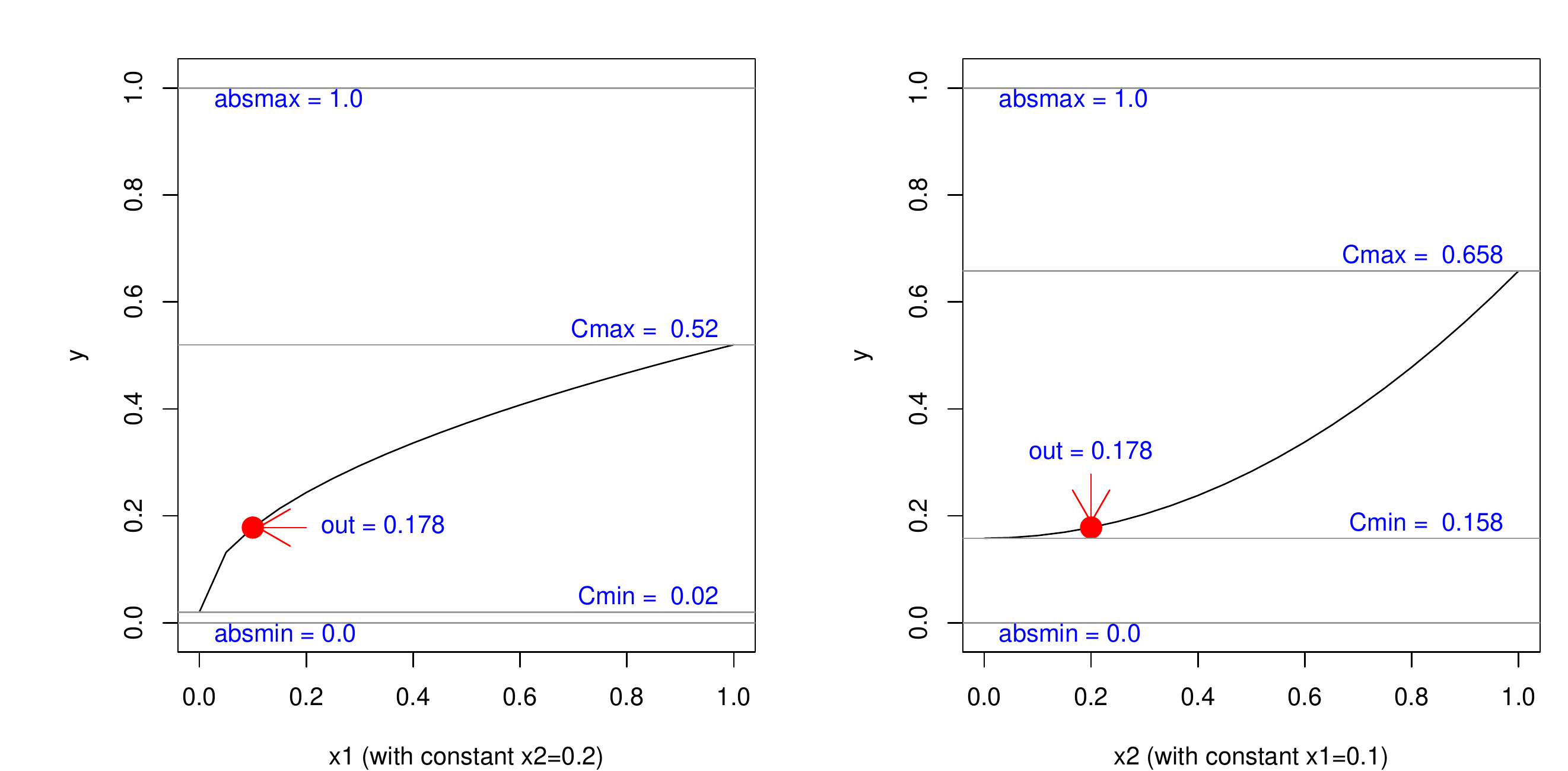}
\caption{Illustration of calculations of CI and CU for the non-linear model in Figure \ref{Fig:nonlinear_func}.}
\label{Fig:CI_CU_simple}
\end{figure}

We are now ready to provide the first definition of Contextual Importance, using a \textit{Pre-defined output range}, followed by the definition of Contextual Utility. 
\begin{definition}[Contextual Importance]
Contextual Importance $CI_{j}(\vv{C},\{i\})$ is a numeric value that expresses to what extent variations in one or several inputs $\{i\}$ affect the value of an output $j$ of a black-box model $f$, according to
\begin{equation}
CI_{j}(\vv{C},\{i\})=\frac{Cmax_{j}(\vv{C},\{i\})-Cmin_{j}(\vv{C},\{i\})}{absmax_{j}-absmin_{j}} 
\label{Eq:CI}
\end{equation}
\end{definition}

\begin{definition}[Contextual Utility]
Contextual Utility $CU_{j}(\vv{C},\{i\})$ is a numeric value that expresses to what extent the current input values $\vv{C}$ are favorable for the output $y_{j}(\vv{C})$ of a black-box model, according to
\begin{equation}
CU_{j}(\vv{C},\{i\})=\frac{y_{j}(\vv{C})-Cmin_{j}(\vv{C},\{i\})}{Cmax_{j}(\vv{C},\{i\})-Cmin_{j}(\vv{C},\{i\})} 
\label{Eq:CU}
\end{equation}
\end{definition}

CI and CU are illustrated in Figure \ref{Fig:CI_CU_simple} for the non-linear function in Figure \ref{Fig:nonlinear_func}. With $(\vv{C})=(0.1,0.2)$, $CI_{1}(\vv{C},\{1\})=0.5$ and $CI_{1}(\vv{C},\{2\})=0.5$, which signifies that both inputs are exactly as important for the output value. For the utilities, $CU_{1}(\vv{C},\{1\})=0,316$ and $CU_{1}(\vv{C},\{2\})=0.04$, so even though the $x_2$ value is higher than the $x_1$ value, the utility of the $x_1$ value is higher than the utility of the $x_2$ value for the result $y$. 

The estimation of the range $[Cmin_{j}(\vv{C},\{i\}),Cmax_{j}(\vv{C},\{i\})]$ is the only part of CIU that requires more than one $\vv{y}=f(\vv{x})$ calculation. It is also the most critical part of CIU for producing explanations that truly correspond to and explain the behaviour of the black-box. Even though it might be possible to calculate or estimate $[Cmin_{j}(\vv{C},\{i\}),Cmax_{j}(\vv{C},\{i\})]$ directly for some models, that is not the case for generic black-box models. One possible approach is to generate a \textit{Set of representative input vectors}. 

\begin{definition}[Set of representative input vectors]
$S(\vv{C},\{i\})$ is an $N \times M$ matrix, where $M$ is the length of $\vv{x}$ and $N$ is a parameter that gives the number of input vectors to generate for obtaining an adequate estimate of the \textit{Estimated output range} $[Cmin_{j}(\vv{C},\{i\}),Cmax_{j}(\vv{C},\{i\})]$.
\end{definition}

A simple way to construct $S(\vv{C},\{i\})$ is to set all input vectors in $S(\vv{C},\{i\})$ to $\vv{C}$ and then replace the values of inputs $\{i\}$ with random values from a pre-defined value range that may be different for every input $x_i$. $N$ is the only adjustable parameter of CIU and needs to be determined based on the complexity of the function learned by the model. More efficient approaches than random values certainly exist but that remains a topic of future research. Furthermore, random values do not guarantee that the generated input vectors are \textit{`representative'}. It might even result in input vectors that are impossible in reality. There is also a risk to have input vectors that are not even close to the examples that were included in the training set of the black-box. This challenge can be addressed at least in the following ways: 

\begin{enumerate}
    \item Use a black-box model that has some guarantees that  $[Cmin_{j}(\vv{C},\{i\}),Cmax_{j}(\vv{C},\{i\})]$ does not go out-of-bounds even with `non-representative' input vectors. In a classification task, for instance, `non-representative' input vectors are not a problem for models whose outputs do not go under zero and do not go over one under any conditions .
    \item Eliminate or correct input vectors that are impossible in reality or that are too far from those included in the training set. One way of doing this could be to remove all rows in $S(\vv{C},\{i\})$ that are too far from any example in the training set. One example of non-realistic input vectors that are straightforward to correct is if there are one-hot encoded inputs, where only one of the concerned inputs is allowed to be TRUE in every input vector. 
    \item Use `non-representative' input vectors on purpose for potentially detecting inconsistencies in the learned model. 
\end{enumerate}

Now that we have studied how to estimate CI and CU of one or more inputs $\{i\}$ for any output  $out_{j}$, we will introduce the notion of \textit{Intermediate Concept}. 

\begin{definition}[Intermediate Concept]
An Intermediate Concept names a given set of inputs $\{i\}$. 
\end{definition}

As defined by Equations (\ref{Eq:CI}) and (\ref{Eq:CU}), CIU can be estimated for any set of inputs $\{i\}$. Intermediate concepts make it possible to specify vocabularies that can be used for producing explanations on any level of abstraction. Different names can be used for the same Intermediate Concept (as well as for input features) and change the concept name used according to the current context and the target explainee(s). 

In addition to using Intermediate Concepts for explaining $y$ values, Intermediate Concept values can be explained using more specific Intermediate Concepts or input features. The following defines \textit{Generalized Contextual Importance} for explaining Intermediate Concepts. 

\begin{definition}[Generalized Contextual Importance]
\begin{equation}
CI_{j}(\vv{C},\{i\},\{I\})=\frac{Cmax_{j}(\vv{C},\{i\})-Cmin_{j}(\vv{C},\{i\})}{Cmax_{j}(\vv{C},\{I\})-Cmin_{j}(\vv{C},\{I\})} 
\label{Eq:CIIntermediateconcept}
\end{equation} 
where $\{I\}$ is the set of input indices that correspond to the Intermediate Concept that we want to explain and $\{i\} \in \{I\}$.
\end{definition}

Equation \ref{Eq:CIIntermediateconcept} is similar to Equation \ref{Eq:CI} when $\{I\}$ is the set of all inputs, i.e. the range $[absmin_{j},absmax_{j}]$ has been replaced by the range $[Cmin_{j}(\vv{C},\{I\},Cmax_{j}(\vv{C},\{I\}]$. Equation \ref{Eq:CU} for CU does not change by the introduction of Intermediate Concepts. In other words, Equation \ref{Eq:CIIntermediateconcept} allows the explanation of the outputs $y_{j}$ as well as the explanation of any Intermediate Concept that leads to $y_{j}$. 



\section{Experimental Results}\label{Sec:Experiments}

Experimental results are shown for two well-know benchmark data sets: Iris flowers and Boston Housing. Iris flowers is a classification task, whereas Boston Housing is a regression task. This choice of using simple but well-known data sets signifies that it is relatively easy to understand the learned models and also to assess what `correct' explanations might look like. 

Bar plot visualisations are used in this paper for illustrating CIU. The length on the bar corresponds to the CI value. A configurable threshold value of $CU_{neutral}=0.5$ has been used for dividing the CU range $[0,1]$ into `defavorable' and `favorable' ranges. A red-yellow-green color scale visualises the CU value, where $CU \in [0,CU_{neutral}]$ gives a continuous transition from red to yellow. $CU \in [CU_{neutral},1]$ gives a continuous transition from yellow to dark green. Future analysis and validation with more data-sets and real-life applications will be needed in order to assess whether $CU_{neutral}$ needs to be adjusted in practice. 

A set $S(\vv{C},\{i\})$ with $N=1000$ has been used for all results reported here, which gives a negligible calculation time using RStudio Version 1.2 on a MacBook Pro from 2017 with a 2,8 GHz Quad-Core Intel Core i7 processor and 16 GB 2133 MHz LPDDR3 memory. 

\subsection{Iris Results}\label{Sec:IrisExperiments}

The neural net described in \cite{Framling_CIU_1995} was used for the Iris data set, which has the useful property of converging towards the mean output value when the input values go towards infinity. Therefore the range $[Cmin_{j}(\vv{C},\{i\}),Cmax_{j}(\vv{C},\{i\})]$ should remain within reasonable bounds. A specific $Iris_{test}$ instance is studied that is quite a typical Virginica with values $\vv{C}=(7,3.2,6,1.8)$ for the input features `Sepal Length', `Sepal Width', `Petal Length', `Petal Width'. The trained neural network gives us $\vv{y}=(0.012, 0.158, 0.830)$ for the three outputs classes `Setosa', `Versicolor', and `Virginica', so it is clearly a Virginica. Table \ref{Tab:CIU_Iris_all} shows the corresponding CIU values for $Iris_{test}$. 

\begin{figure}
\centering
\begin{subfigure}{\textwidth}
\centering
\includegraphics[width=\textwidth]{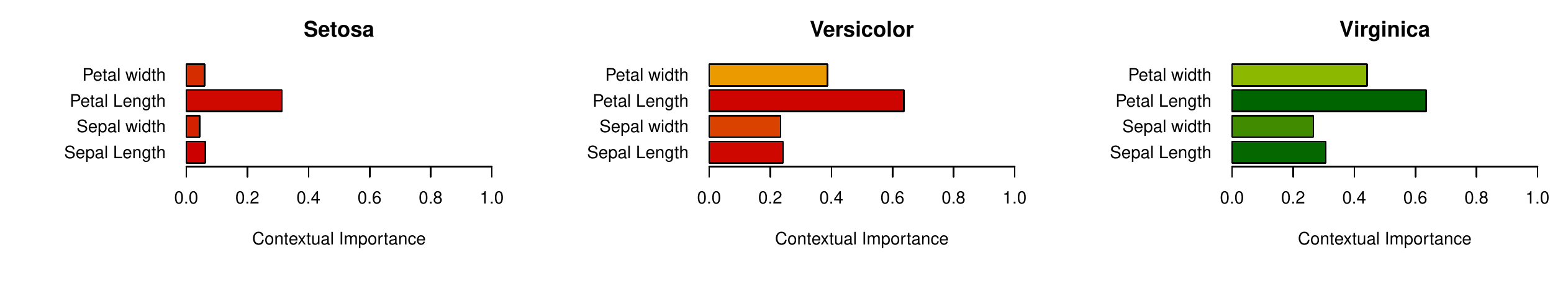}
\caption{All four inputs versus Iris class.} \label{Fig:IrisCIU_barplot_INKA}
\end{subfigure}
\begin{subfigure}{\textwidth}
\centering
\includegraphics[width=\textwidth]{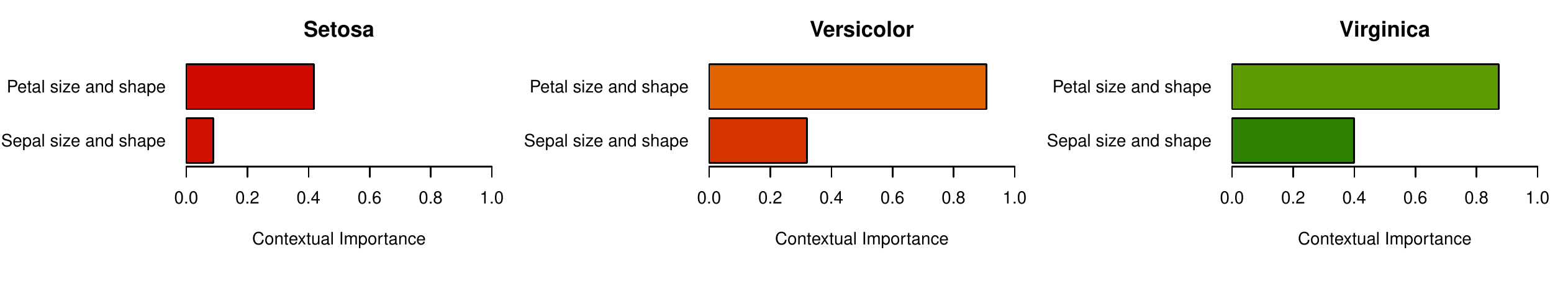}
\caption{Intermediate Concepts `Sepal size and shape' and `Petal size and shape' versus Iris class.}
\label{Fig:IrisCIU_IntermediateConcepts_INKA}
\end{subfigure}
\begin{subfigure}{\textwidth}
\centering
\includegraphics[width=\textwidth]{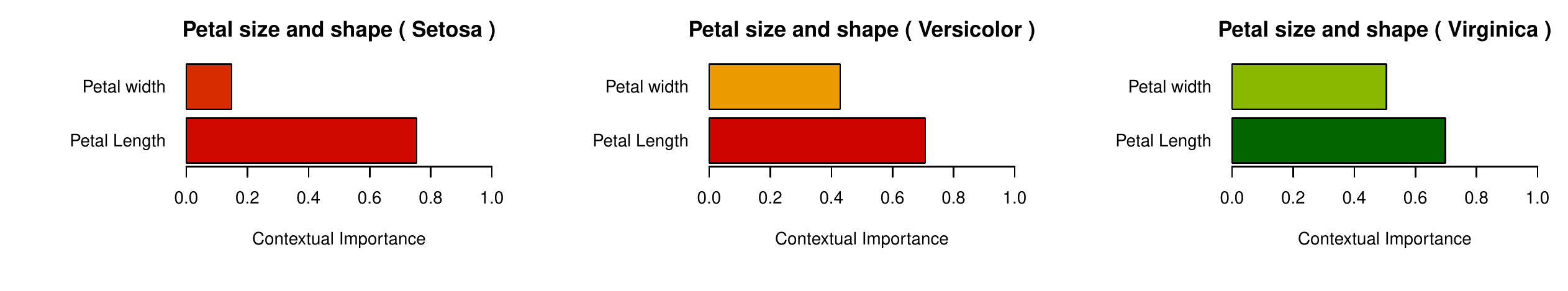}
\caption{CIU of `Petal width' and `Petal length' versus Intermediate Concept `Petal size and shape'.}
\label{Fig:IrisCIU_INKA_IntermediateExplanations}
\end{subfigure}
\caption{CIU bar plot visualisations for Iris task. Bar lengths correspond to CI values. CU values are visualised using a continuous red-yellow-green color palette.}
\label{Fig:IrisBarplots_INKA}
\end{figure}

\begin{figure}
\centering
\begin{subfigure}{\textwidth}
\centering
\includegraphics[width=0.49\textwidth]{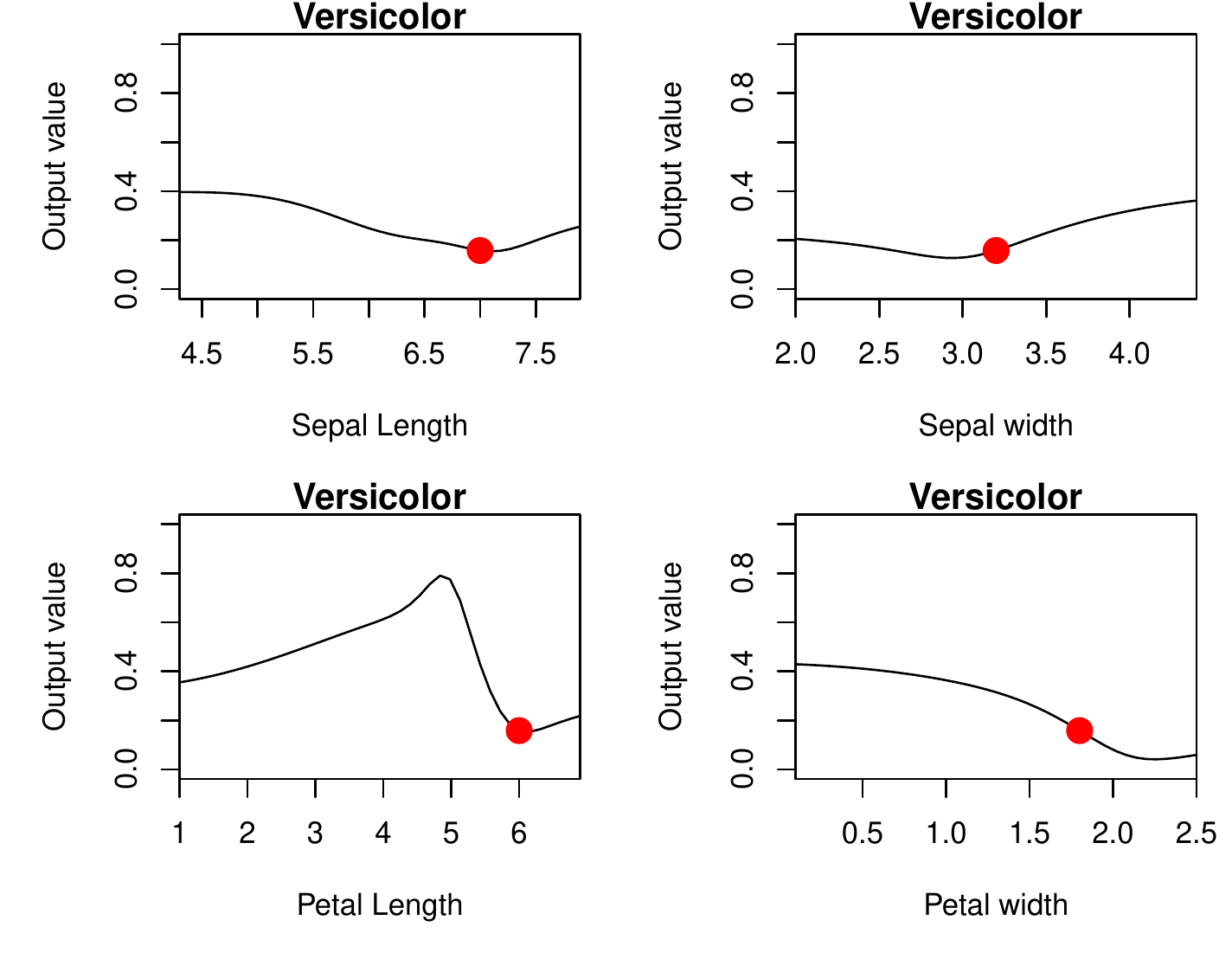}
\includegraphics[width=0.49\textwidth]{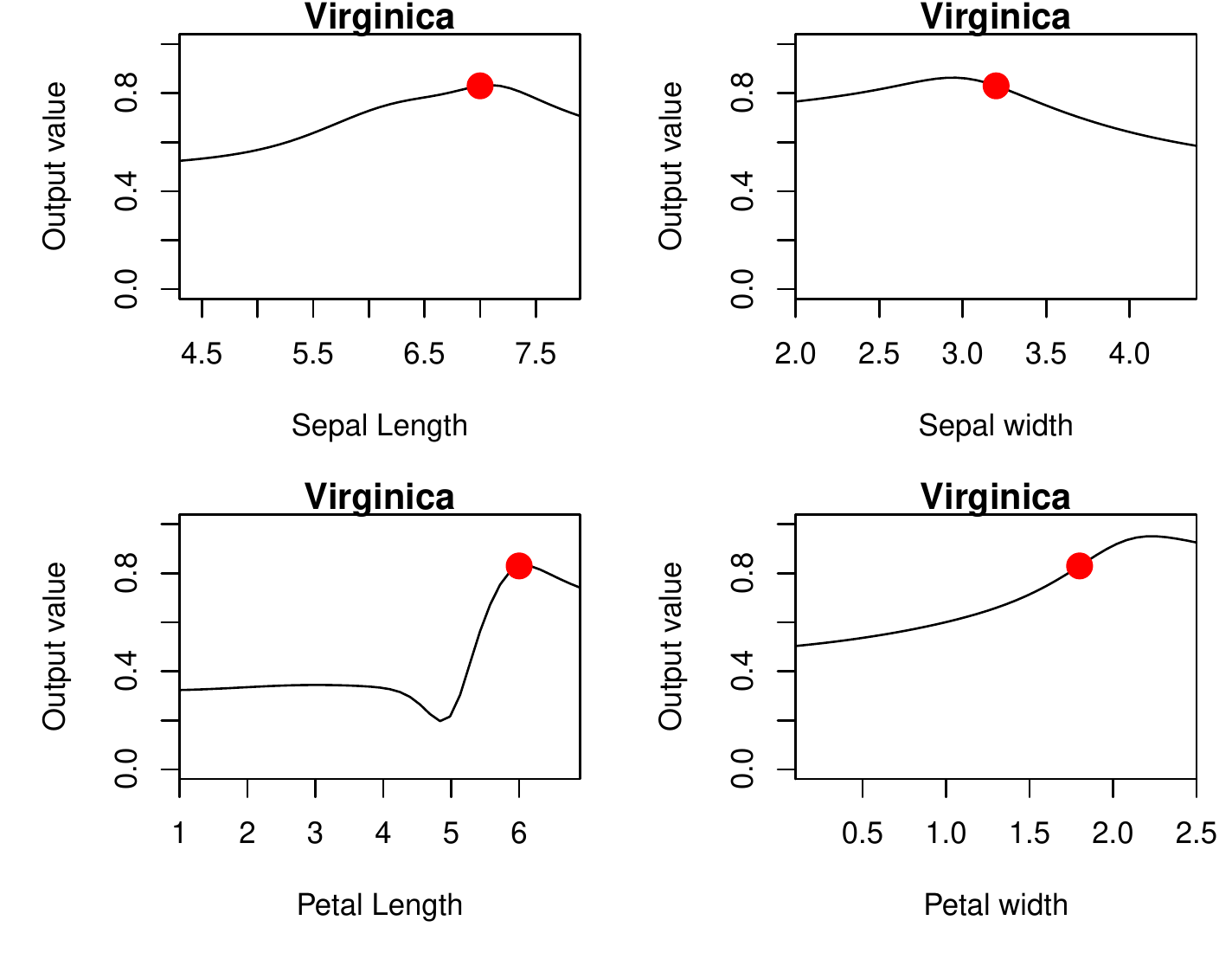}
\label{Fig:IrisCIU_Versicolor_Virginica_INKA}
\end{subfigure}
\begin{subfigure}{\textwidth}
\centering
\includegraphics[trim={1cm 0cm 0cm 0cm},clip,width=0.8\textwidth]{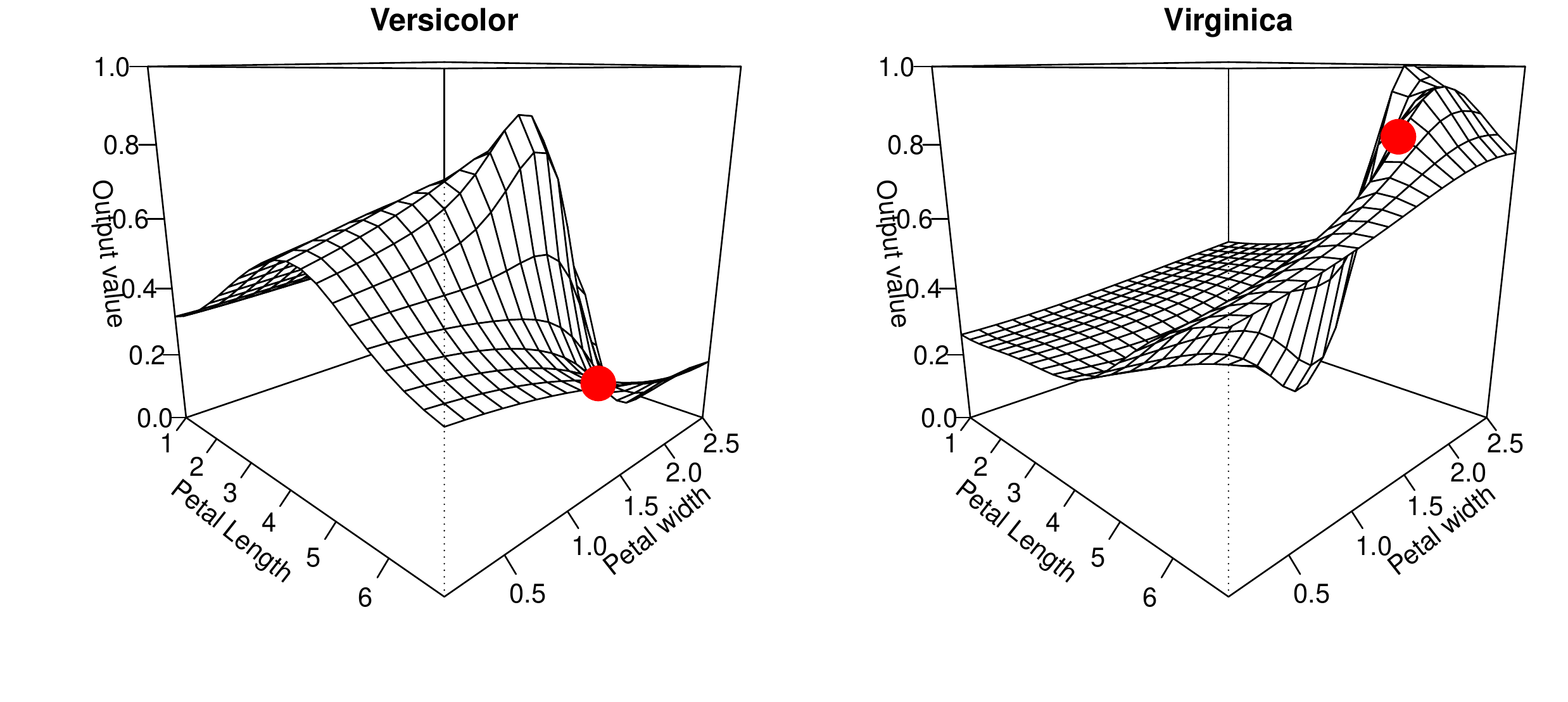}
\label{Fig:IrisCIU_Versicolor_Virginica_INKA_3D}
\end{subfigure}
\caption{Output $y_{j}$ as a function of input values for Versicolor (left) and Virginica (right). Red dot shows input and output values for $Iris_{test}$.}
\label{Fig:IrisCIU_Versicolor_Virginica_INKA_all}
\end{figure}

Some questions that could be asked are `\textit{Why is it a Virginica?}' but also `\textit{Why is it not a Versicolor or a Setosa?}'. Figure \ref{Fig:IrisCIU_barplot_INKA} shows bar plot explanations for the three Iris classes. It is clear that $Iris_{test}$ is not a Setosa because none of the features is typical for a Setosa and modifying any of the values will not change the situation. On the other hand, all features are typical for a Virginica. Petal length is clearly the most important feature for the classification of $Iris_{test}$. Figure \ref{Fig:IrisCIU_Versicolor_Virginica_INKA_all} shows how the output value (estimated class probability) changes for Versicolor and Virginica as a function of the four input features. These graphs confirm that Petal Length is the feature that discriminates Versicolor and Virginica the most from each other. 

For showing the use of Intermediate Concepts, a small vocabulary was developed. The vocabulary specifies that `Sepal size and shape' is the combination of features `Sepal Length' and `Sepal Width'. `Petal size and shape' is the combination of features `Petal Length' and `Petal Width'. When studying the results using the Intermediate Concepts `Sepal size and shape' and `Petal size and shape', we get the bar plot explanation in Figure \ref{Fig:IrisCIU_IntermediateConcepts_INKA}. 

Finally, Figure \ref{Fig:IrisCIU_INKA_IntermediateExplanations} answers questions such as `why is Petal size and shape not so typical for Versicolor?' and `why is Petal size and shape typical for Virginica?'. These bar plots express what can be observed also in the 3D graphs of Figure \ref{Fig:IrisCIU_Versicolor_Virginica_INKA_all}, where we can see that the combination of `Petal Length' and `Petal Width' could be even more typical for Virginica than what it is for $Iris_{test}$.

\begin{table}
\centering
\caption{CIU values for Iris classes versus input.}\label{tab1}
\begin{tabular}{lllllll}
\hline
Iris class ($y_{j}$) & \multicolumn{2}{c}{Setosa (0.012)} & \multicolumn{2}{c}{Versicolor (0.158)} & \multicolumn{2}{c}{Virginica (0.830)}\\
Input feature & \multicolumn{1}{c}{CI} & \multicolumn{1}{c}{CU} & \multicolumn{1}{c}{CI} & \multicolumn{1}{c}{CU} & \multicolumn{1}{c}{CI} & \multicolumn{1}{c}{CU} \\
\hline
Sepal Length & 0.067 & 0.000 & 0.242 & 0.015 & 0.309 & 0.990\\
Sepal Width & 0.044 & 0.130 & 0.234 & 0.130 & 0.278 & 0.880\\
Petal Length & 0.314 & 0.015 & 0.640 & 0.008 & 0.638 & 0.995\\
Petal Width & 0.061 & 0.087 & 0.388 & 0.302 & 0.448 & 0.729\\
\hline
Sepal size and shape & 0.087 & 0.030 & 0.320 & 0.104 & 0.399 & 0.910\\
Petal size and shape & 0.408 & 0.023 & 0.895 & 0.189 & 0.903 & 0.809\\
\hline
All inputs & 0.869 & 0.013 & 0.927 & 0.110 & 0.920 & 0.886\\
\hline
\end{tabular}
\label{Tab:CIU_Iris_all}
\end{table}

\subsection{Boston Housing Results}\label{Sec:BostonExperiments}

\begin{figure}
\centering
\includegraphics[width=\textwidth]{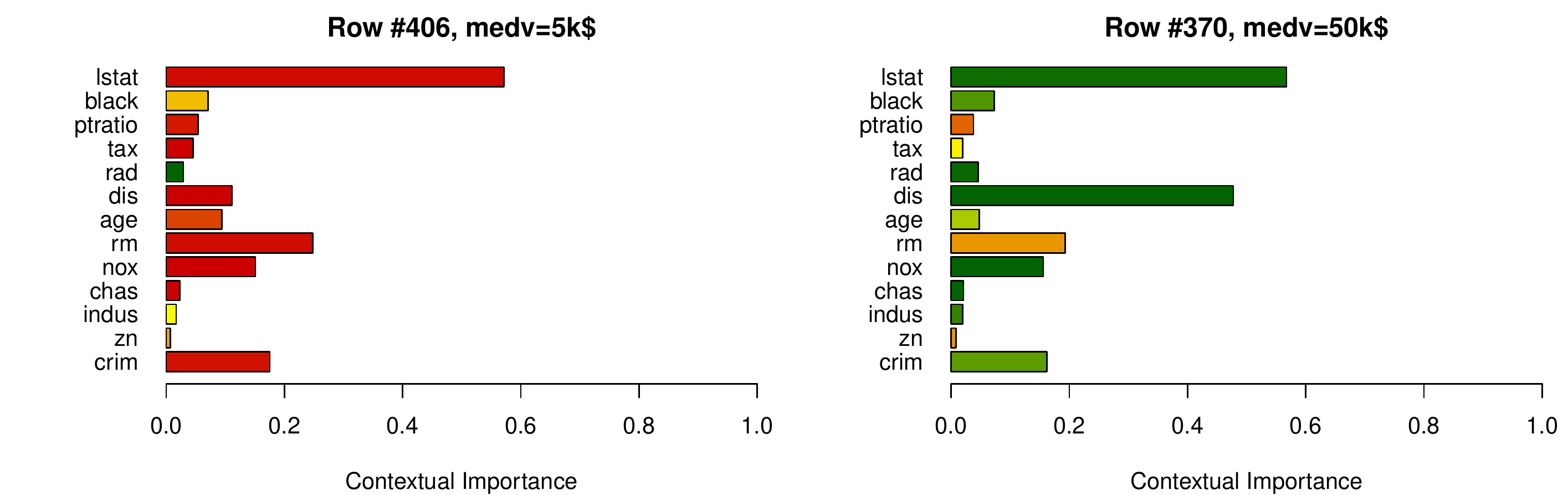}
\caption{CIU for two Boston Housing data set instances.}
\label{Fig:BostonGradientBoostBarplot}
\end{figure}

A gradient boosting model was used for the Boston Housing data set. It learned the mapping from the 13 input variables to the Median value (medv) of owner-occupied homes in \$1000's. The resulting CIU bar plot is shown in Figure \ref{Fig:BostonGradientBoostBarplot} for instances \#406 (medv=5, lowest), and \#370 (medv=50, highest). CIU clearly identifies what are the most important features for the different instances, as well as identifying which input values are favorable for a high `medv' value. These results are consistently similar over different runs. 

When using a neural net like in Section \ref{Sec:IrisExperiments} the learned model also becomes slightly different than with the gradient boosting model. These differences in the underlying black-box model are well reflected also in the CIU explanations, with differences notably in CI values. CU values and the overall balance between red and green in the bar plots remains similar. This illustrates how CIU can also be used as a tool for assessing the credibility and trust in different black-box models. 

\section{Future Work}\label{Sec:Future}

Two simple benchmark data sets were used in this paper for reasons of illustration and to enable human readers to assess the validity of the results. Experience from more data sets and use cases might lead to further extensions of CIU. For instance, use cases involving one-hot coding will require using Intermediate Concepts for aggregating the concerned one-hot inputs into one single explainable feature.  

CIU provides many topics for future research. For instance, is it always better to use CI values as such or normalise them to one? What ways of visualising CIU are the best understood by humans? Research is ongoing in these directions but how to best interact with human explainees is a vast domain. Research has also been initiated for using CIU together with Reinforcement and Unsupervised learning. 

\section{Conclusions}\label{Sec:Conclusions}

CIU is a model-agnostic method that allows producing explanations from any black-box model (no matter how `black' or not it is), without producing an intermediate interpretable model. Therefore CIU does not have the same challenges of black-box model fidelity as most other XAI methods do. Compared to other output explanation methods, CIU allows for more flexibility in how explanations can be produced and presented to explainees due to the possibility to apply CIU to sets of features and Intermediate Concepts. Intermediate Concepts enable the use of different vocabularies depending on the context and on the explainee. The Contextual Utility concept also allows to produce explanations in a more human-like way than other XAI methods. 

By not using an intermediate interpretable model, CIU does not fit into any of the existing categories presented by major XAI survey articles. CIU has only one adjustable parameter, i.e. the number of samples in $S(\vv{C},\{i\})$, which might be possible to eliminate or automate in the future. Therefore, CIU establishes a new category of XAI methods that will hopefully help to solve at least some of the many challenges that AI and XAI are currently facing. 



\bibliographystyle{unsrt}  
\bibliography{KaryBib}

\end{document}